\def\year{2022}\relax

\documentclass[twocolumn,letterpaper]{article}

\usepackage{aaai22, times, helvet, courier, graphicx, natbib, caption, newfloat, listings, bibentry}
\usepackage{enumitem, graphicx, subfigure, multirow, amsfonts, amsmath, xcolor, verbatim, algorithm, algorithmicx, algpseudocode, booktabs, arydshln, times, epsfig, amssymb, multirow, capt-of, pifont, color, xcolor, xspace, diagbox}

\DeclareCaptionStyle{ruled}{labelfont=normalfont,labelsep=colon,strut=off}

\floatstyle{ruled}
\newfloat{listing}{tb}{lst}{}
\floatname{listing}{Listing}

\newcommand\blfootnote[1]{
    \begingroup
    \renewcommand\thefootnote{}\footnote{#1}
    \addtocounter{footnote}{-1}
    \endgroup
}
\newcommand*{\docppt}{\texttt{DOC2PPT}\xspace}

\title{DOC2PPT: Automatic Presentation Slides Generation from Scientific Documents}
\author{
    Tsu-Jui Fu\textsuperscript{\rm 1}, William Yang Wang\textsuperscript{\rm 1}, Daniel McDuff\textsuperscript{\rm 2}, Yale Song\textsuperscript{\rm 2}
}
\affiliations{
    \textsuperscript{\rm 1} UC Santa Barbara~~~\textsuperscript{\rm 2} Microsoft Research \\
    \{tsu-juifu,william\}@cs.ucsb.edu~~~\{damcduff, yalesong\}@microsoft.com
}

\begin{document}
\maketitle

\begin{abstract}
Creating presentation materials requires complex multimodal reasoning skills to summarize key concepts and arrange them in a logical and visually pleasing manner. Can machines learn to emulate this laborious process? We present a novel task and approach for document-to-slide generation. Solving this involves document summarization, image and text retrieval, and slide structure to arrange key elements in a form suitable for presentation. We propose a hierarchical sequence-to-sequence approach to tackle our task in an end-to-end manner. Our approach exploits the inherent structures within documents and slides and incorporates paraphrasing and layout prediction modules to generate slides. To help accelerate research in this domain, we release a dataset of about 6K paired documents and slide decks used in our experiments. We show that our approach outperforms strong baselines and produces slides with rich content and aligned imagery. \blfootnote{Project webpage: https://doc2ppt.github.io/}
\end{abstract}

\section{Introduction}
Creating presentations is often a work of art. It requires skills to abstract complex concepts and conveys them in a concise and visually pleasing manner. Consider the steps involved in creating presentation slides based on a white paper or manuscript: One needs to 1) establish a storyline that will connect with the audience, 2) identify essential sections and components that support the main message, 3) delineate the structure of that content, e.g., the ordering/length of the sections, 4) summarize the content in a concise form, e.g., punchy bullet points, and 5) gather figures that help communicate the message accurately and engagingly.

\begin{figure}[!t]
\centering
    \includegraphics[width=.8\linewidth]{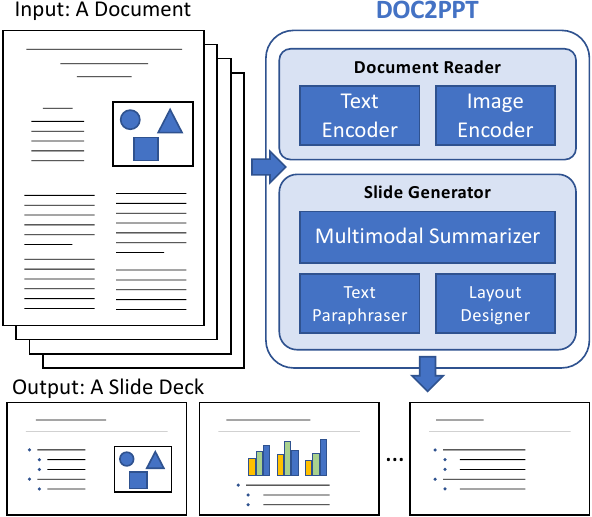}
    \caption{We introduce \docppt, a novel task of generating a slide deck from a document. This requires solving several challenges in the vision-and-language domain, e.g., visual-semantic embedding and multimodal summarization. In addition, slides exhibit unique properties such as concise text (bullet points) and stylized layout.}
    \label{fig:concept}
    \vspace{-3ex}
\end{figure}

Can machines emulate this laborious process by \textit{learning} from the plethora of example manuscripts and slide decks created by human experts? Building such a system poses unique challenges in vision-and-language understanding. Both the input (a manuscript) and output (a slide deck) contain tightly coupled visual and textual elements; thus, it requires multimodal reasoning. Further, there are significant differences in the presentation: compared to manuscripts, slides tend to be more \textit{concise} (e.g., containing bullet points rather than full sentences), \textit{structured} (e.g., each slide has a fixed screen real estate and delivers one or few messages), and \textit{visual-centric} (e.g., figures are first-class citizens, the visual layout plays an important role, etc.).

Existing literature only partially addresses some of the challenges above. Document summarization~\cite{cheng16ext-summ,chopra16seq2seq-summ} aims to find a concise text summary of the input, but it does not deal with images/figures and lacks multimodal understanding. Cross-modal retrieval~\cite{frome13devise,kiros14text-image-emb} focuses on finding a multimodal embedding space but does not produce summarized outputs. Multimodal summarization~\cite{zhu18msmo} deals with both (summarizing documents with text and figures), but it lacks the ability to produce structured output (as in slides). Furthermore, none of the above addresses the challenge of finding an optimal visual layout of each slide. While assessing visual aesthetics have been investigated~\cite{marchesotti11assessing}, exiting work focuses on photographic metrics for images that would not translate to slides. These aspects make ours a unique task in the vision-and-language literature.

In this paper, we introduce \docppt, a novel task of creating presentation slides from scientific documents. As with no existing benchmark, we collect 5,873 paired scientific documents and associated presentation slide decks (for a total of about 70K pages and 100K slides, respectively). We present a series of automatic data processing steps to extract useful learning signals and introduce new quantitative metrics designed to measure the quality of the generated slides.

To tackle this task, we present a hierarchical recurrent sequence-to-sequence architecture that ``reads'' the input document and ``summarizes'' it into a \textit{structured} slide deck. We exploit the inherent structure within documents and slides by performing inference at the section-level (for documents) and at the slide-level (for slides). To make our model end-to-end trainable, we explicitly encode section/slide embeddings and use them to learn a policy that determines \textit{when to proceed} to the next section/slide. Further, we learn the policy in a hierarchical manner so that the network decides which actions to take by considering the structural context, e.g., a decision to create a new slide will depend on both the current section and the previous generated content. 

To consider the concise nature of text in slides (e.g., bullet points), we incorporate a paraphrasing module that converts document-style full sentences to slide-style phrases/clauses. We show that it drastically improves the quality of the generated textual content for the slides. In addition, we introduce a text-figure matching objective that encourages related text-figure pairs to appear on the same slide. Lastly, we explore both template-based and learning-based layout design and compare them both quantitatively and qualitatively.

Taking a long-term view, our objective is not to take humans completely out of the loop but enhance humans' productivity by generating slides \textit{as drafts}. This would create new opportunities to human-AI collaboration~\cite{amershi2019guidelines}, e.g., one could quickly create a slide deck by revising the auto-generated draft and skim them through to digest lots of material. To summarize, our main contributions include: 1) Introducing a novel task, dataset, and evaluation metrics for automatic slide generation; 2) Proposing a hierarchical sequence-to-sequence approach that summarizes a document in a structure output format suitable for slide presentation; 3) Evaluating our approach both quantitatively, using our proposed metrics, and qualitatively based on human evaluation. We hope that our \docppt will advance the state-of-the-art in the vision-and-language domain.

\section{Related Work}
\paragraph{Vision-and-Language.} Joint modeling of vision-and-language has been studied from different angles. Image/video captioning~\cite{vinyals16show,you16image,li16tgif,xu16msr}, visual question answering~\cite{antol15vqa,jang17tgif,anderson18bottom}, and visually-grounded dialogue generation~\cite{das17visual} are all tasks that involve learning relationships between image and text. Despite this large body of work, there remain many tasks that have not been addressed, e.g., multimodal document generation. As argued above, our task brings a new suite of challenges to vision-and-language understanding.

\paragraph{Document Summarization.} This task has been tackled from two angles: abstractive~\cite{chopra16seq2seq-summ,see17ptr-summ,cho19selector-summ,liu19bert-summ,dong19unilm,zhang20pegasus,celikyilmaz18dca,rush15abs-summ,liu18gan-summ,paulus18rl-summ} and extractive~\cite{barrios15textrank-summ,narayan18refresh,liu19bert-ext-summ,chen18its,yin14sel-summ,cheng16ext-summ,yasunaga17graph-summ}. Our \docppt task involves both abstractive and extractive summarization since it requires to extract the key content from a document \textit{and} paraphrase it into a concise form. A task closely related to ours is scientific document summarization \cite{elkiss08cite-summ,lloret13compendium,hu2013ppsgen,jaidka16clscisumm,parveen16cp-scisumm,sefid2019automatic}, but to date that work has only focused on producing text summaries, while we focus on generating multimedia slides. Furthermore, existing datasets in this domain (such as TalkSumm~\cite{lev19talksumm} and ScisummNet~\cite{yasunaga19scisummnet}) are rather small with only about 1K documents each. We propose a large dataset of 5,873 pairs of high-quality scientific documents and slide decks.

\paragraph{Visual-Semantic Embedding.} Our task involves generating slides with relevant text and figures. Learning text-image similarity has been studied in the visual-semantic embedding (VSE) literature~\cite{karpathy14text-image,vendrov16text-image-emb,faghri18vse,huang18text-image,gu18text-image-gan,song19polysemous}. However, unlike the VSE setting where text instances are known in advance, ours requires simultaneously \textit{generating} text and retrieving the related images at the same time.

\paragraph{Multimodal Summarization.} MSMO \cite{zhu18msmo,zhu20msmo,li2019vmsmo} generates textual summarization with related images for news articles. Similarly, our task includes summarizing multimodal documents, but it also involves putting the summary in a structured format such as slides. 

\section{Approach}
The goal of \docppt is to generate a slide deck from a multimodal document with text and figures.\footnote{In this work, figures include images, graphs, charts, and tables.} As shown in Fig.~\ref{fig:concept}, the task involves ``reading'' a documen and summarizing it, paraphrasing the summarized sentences into a concise format suitable for slide presentation, and placing the chosen text and figures to appropriate locations in the output slides.

\begin{figure}[t]
\centering
    \includegraphics[width=.9\linewidth]{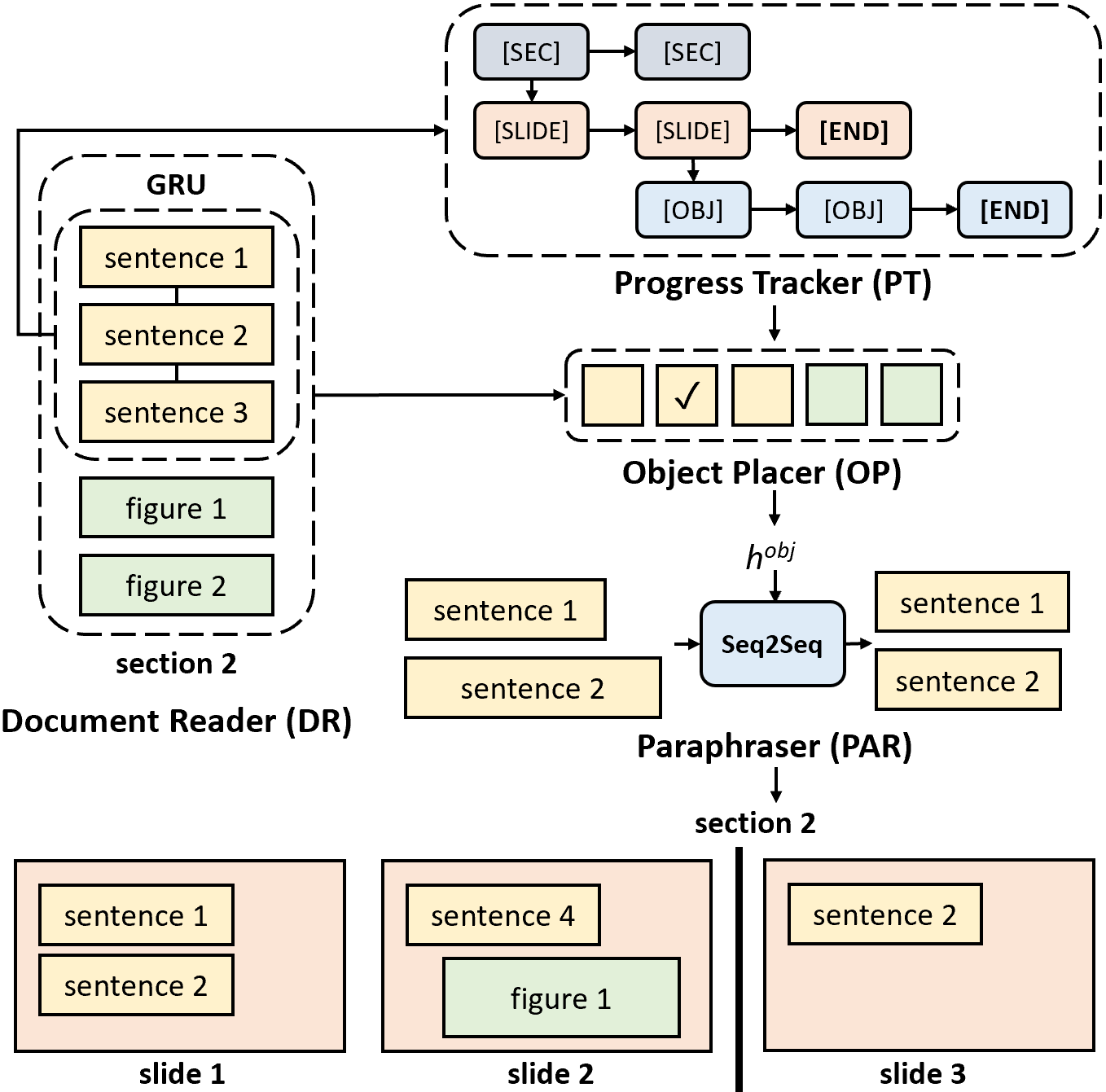}
    \caption{An overview of our architecture. It consists of modules (DR, PT, OP, PAR) that read a document and generate a slide deck in a hierarchically structured manner.}
    \label{fig:overall}
    \vspace{-3ex}
\end{figure}

\paragraph{Overview.} Given the multi-objective nature of the task, we design our network with modularized components that are jointly trained in an end-to-end fashion. Fig.~\ref{fig:overall} shows an overview of our network that includes these modules:
\setlist{nolistsep}
\begin{itemize}[noitemsep]
\item A \textbf{Document Reader (DR)} encodes sentences and figures in a document;
\item A \textbf{Progress Tracker (PT)} maintains pointers to the input (i.e., which section is currently being processed) and the output (i.e., which slide is currently being generated) and determines when to proceed to the next section/slide based on the progress so far;
\item An \textbf{Object Placer (OP)} decides which object from the current section (sentence or figure) to put on the current slide. It also predicts the location and the size of each object to be placed on the slide;
\item A \textbf{Paraphraser (PAR)} takes the selected sentence and rewrites it in a concise form before putting it on a slide.
\end{itemize}

\paragraph{Notation.} A document $\mathcal{D}$ is organized into sections $\mathcal{S}=\{S_i\}_{i \in N^{in}_S}$ and figures $\mathcal{F}=\{F^{in}_q\}_{q \in M^{in}_F}$. Each section $S_i$ contains sentences $\mathcal{T}^{in}_i=\{T^{in}_{i, k}\}_{k \in N^{in}_i}$, and each figure $F_q=\{I_q, C_q\}$ contains an image $I_q$ and a caption $C_q$. We do not assign figures to any particular section because multiple sections can reference the same figure. A slide deck $\mathcal{O}=\{O_j\}_{j \in N^{out}_O}$ contains a number of slides, each containing sentences $\mathcal{T}^{out}_j=\{T^{out}_{j,k}\}_{k \in N^{out}_j}$ and figures $\mathcal{F}^{out}_j=\{F^{out}_{j, k}\}_{k \in M^{out}_j}$. We encode the position and the size of each object on a slide in a bounding box format using an auxiliary layout variable $L_{j, k}$, which includes four real-valued numbers $\{l^x, l^y, l^w, l^h\}$ encoding the x-y offsets (top-left corner), the width and height of a bounding box.

\subsection{Model}
\paragraph{Document Reader (DR).} We extract sentence and figure embeddings from an input document and project them to a shared embedding space so that the OP treats both textual and visual elements as an object coming from a joint multimodal distribution. For each section $S_i$, we use RoBERTa~\cite{liu19roberta} to encode each of the sentences $T^{in}_{i, k}$, and then use a bidirectional GRU \cite{chung14gru} to extract contextualized sentence embeddings $X^{in}_{i, k}$:
\begin{equation}
\begin{split}
    B^{in}_{i, k} &= \text{RoBERTa}(T^{in}_{i, k}), \\
    X^{in}_{i, k} &= \text{Bi-GRU}{(B^{in}_{i, 0}, ..., B^{in}_{i, N^{in}_i-1})}_{k},
\end{split}
\end{equation}
Similarly, for each figure $F^{in}_{q}=\{I^{in}_q, C^{in}_q\}$, we apply ResNet-152~\cite{he16resnet} to extract the image embedding of $I^{in}_q$ and RoBERTa for the caption embedding of $C^{in}_{q}$. We then concatenate them as the figure embedding $V^{in}_{q}$:
\begin{equation}
\begin{split}
    V^{in}_{q} &= [\text{ResNet}(F^{in}_{q}), \text{RoBERTa}(C^{in}_{q})].
\end{split}
\end{equation}

Next, we project $X^{in}_{i, k}$ and $V^{in}_{q}$ to a shared embedding using a two-layer multilayer perceptron (MLP) and combine $E^{txt}_{i}$ and $E^{fig}$ as the section embedding $E^{sec}_{i}$ of $S_i$:
\begin{equation}
\begin{split}
\label{eq:emb-e}
    E^{txt}_{i, k} = \text{MLP}^{txt}(X^{in}_{i, k}), \:\:\:\: E^{fig}_{q} = \text{MLP}^{fig}(V^{in}_{q}), \\
    E^{sec}_{i} = \{ E^{txt}_{i, k}, E^{fig}_{q} \}_{k \in N^{in}_{i}, q \in M^{in}_{F}}
\end{split}
\end{equation}
We include all figures $\mathcal{F}$ in \textit{each} section embedding $E^{sec}_{i}$ because each section can reference any of the figures.

\paragraph{Progress Tracker (PT).} We define the PT as a state machine operating in a hierarchically-structured space with sections ([SEC]), slides ([SLIDE]), and objects ([OBJ]). This is to reflect the structure of documents and slides, i.e., each section of a document can have multiple corresponding slides, and each slide can contain multiple objects. 

The PT maintains pointers to the current section $i$ and the current slide $j$, and learns a policy to proceed to the next section/slide as it generates slides. For simplicity, we initialize $i=j=0$, i.e., the output slides will follow the natural order of sections in an input document. We construct PT as a three-layer hierarchical RNN with $( \texttt{PT}^{sec},\texttt{PT}^{slide},\texttt{PT}^{obj} )$, where each RNN encodes the latent space for each level in a section-slide-object hierarchy. This is a natural choice to encode our prior knowledge about the hierarchical structure.

First, $\texttt{PT}^{sec}$ takes as input the head-tail contextualized sentence embeddings from the DR, which encodes the overall information of the current section $S_i$. We use GRU for $\texttt{PT}^{sec}$ and initialize $h^{sec}_{0}$ to the contextualized sentence embeddings of the first section, i.e., $h^{sec}_{0}=[X^{in}_{0, 1}, X^{in}_{0, N^{in}_0 - 1}]$:
\begin{equation}
    h^{sec}_{i} = \texttt{PT}^{sec}(h^{sec}_{i-1}, [X^{in}_{i, 1}, X^{in}_{i, N^{in}_i}]),
\end{equation}

Based on the section state $h^{sec}_{i}$, $\texttt{PT}^{slide}$ models the section-to-slide relationships: 
\begin{equation}
    a^{sec}_j, h^{slide}_{j} = \texttt{PT}^{slide}(a^{sec}_{j-1}, h^{slide}_{j-1}, E^{sec}_{i}),
\end{equation}
where $h^{slide}_{0} = h^{sec}_{i}$, $E^{sec}_{i}$ is the section embedding (Eq.~\ref{eq:emb-e}), and $a^{sec}_j$ is a binary action variable that tracks the section pointer, i.e, it decides if the model should generate a new slide for the current section $S_i$ or proceed to the next section $S_{i+1}$. We implement $\texttt{PT}^{slide}$ as a GRU and a two-layer MLP with a binary decision head that learns a policy $\phi$ to predict $a^{sec}_j = \{ \texttt{[NEW\_SLIDE]}, \texttt{[END\_SEC]}\}$:
\begin{equation}
\begin{split}
    a^{sec}_j &= \text{MLP}^{slide}_{\phi}([h^{slide}_j, \sum\nolimits_r \alpha^{slide}_{j, r} E^{sec}_{i, r}]), \\
    \alpha^{slide}_j &= \text{softmax}(h^{slide}_j W (E^{sec}_i)^{\intercal}).
\end{split}
\end{equation}
$\alpha^{slide}_j \in \mathbb{R}^{N_i^{in}+M^{in}}$ is an attention map over $E^{sec}_i$ that computes the bilinear compatibility between $h^{slide}_j$ and $E^{sec}_i$.

Finally, the object $\texttt{PT}^{obj}$ tracks which objects to put on the current slide $O_j$ based on the slide state $h^{slide}_{j}$:
\begin{equation}
\begin{split}
    a^{slide}_k, h^{obj}_{k} &= \texttt{PT}^{obj}(a^{slide}_{k-1}, h^{obj}_{k-1}, E^{sec}_{i}), \\
    a^{slide}_k &= \text{MLP}^{obj}_{\psi}([h^{obj}_k, \sum\nolimits_r \alpha^{obj}_{k, r} E^{sec}_{i, r}]), \\
    \alpha^{obj}_k &= \text{softmax}(h^{obj}_k W (E^{sec}_i)^{\intercal}).
\end{split}
\end{equation}
Similarly, $a^{slide}_k= \{ \texttt{[NEW\_OBJ]},$ $\texttt{[END\_SLIDE]}\}$ is a binary action variable that decides whether to put a new object for the current slide or proceed to the next. We again set $h^{obj}_{0} = h^{slide}_{j}$ and use a GRU and a two-layer MLP$_\psi$ to implement $\texttt{PT}^{obj}$, together with an attention matrix $W$ between $h^{obj}_k$ and $E^{sec}_i$. Note that each of the three PTs have an independent set of weights to ensure that they model distinctive dynamics in the section-slide-object structure.

\paragraph{Object Placer (OP).} When $\texttt{PT}^{obj}$ takes an action $a^{slide}_k=\texttt{[NEW\_OBJ]}$, the OP selects an object from the current section $S_i$ and predicts the location on the current slide $O_j$ in which to place it. For this, we use the attention score $\alpha^{obj}_k$ to choose an object (sentence or figure) that has the maximum compatibility score with the current object state $h^{obj}_k$, i.e., $\arg\max_r \alpha^{obj}_k$. We then employ a two-layer MLP to predict the layout variable for the chosen object:
\begin{equation}
    \{l^x_k, l^y_k, l^w_k, l^h_k\} = \text{MLP}^{layout}([h^{obj}_k, \sum\nolimits_r \alpha^{obj}_{k, r} E^{sec}_{i, r}]).
\end{equation}

Note that the distinctive style of presentation slides requires special treatment of the objects. If an object is a figure, we take only the image part and resize it to fit the bounding box region while maintaining the original aspect ratio. If an object is a sentence, we first paraphrase it into a concise form and also adjust the font size to fit inside.

\paragraph{Paraphraser (PAR).} We paraphrase sentences before placing them on slides. This step is crucial because without it the text would be too verbose for a slide presentation.\footnote{In our dataset, sentences in the documents have an average of 17.3 words, while sentences in slides have 11.6 words; the difference is statistically significant ($p=0.0031$).} We implement the PAR as Seq2Seq~\cite{bahdanau15seq2seq} with the copy mechanism~\cite{gu16copy}:
\begin{equation}
    \{w_0, ..., w_{l-1}\} = \text{PAR}(T_{j,k}^{out}, h_k^{obj}),
\end{equation}
where $T_{j,k}^{out}$ is a sentence chosen by OP. We condition PAR on the object state $h_{k}^{obj}$ to provide contextual information and demonstrate this importance in the supplementary.

\begin{table*}[tp]
\centering
\footnotesize
    \begin{tabular}[t]{lccrrrrrrr}
        \toprule
        & \multicolumn{1}{c}{\textbf{Document - Slide}} & ~ & \multicolumn{3}{c}{\textbf{Documents}} & ~ & \multicolumn{3}{c}{\textbf{Slides}} \\
        \cmidrule{2-2} \cmidrule{4-6} \cmidrule{8-10} 
        ~ & Train / Val / Test & ~ & \#Sections & \#Sentences & \#Figures & ~ & \#Slides & \#Sentences & \#Figures \\
        \midrule
        CV & 2,073 / 265 / 262 & ~ & 15,588 (6.0) & 721,048 (46.3) & 24,998 (9.6) & ~ & 37,969 (14.6) & 124,924 (8.0) & 4,290 (1.7) \\
        NLP  & $\:\:\:$741 / $\:\:$93 / $\:\:$97 & ~ & 7,743 (8.3) & 234,764 (30.3) & 8,114 (8.7) & ~ & 19,333 (20.8) & 63,162 (8.2) & 3,956 (4.2) \\
        ML & 1,872 / 234 / 236 & ~ & 17,735 (7.6) & 801,754 (45.2) & 15,687 (6.7) & ~ & 41,544 (17.7) & 142,698 (8.0) & 6,187 (2.6) \\ \midrule
        Total & 4,686 / 592 / 595 & ~ & 41,066 (6.99) & 1,757,566 (42.8) & 48,799 (8.3) & ~ & 98,856 (16.8) & 330,784 (8.1) & 14,433 (2.5) \\
        \bottomrule
    \end{tabular}
    \caption{Descriptive statistics of our dataset. We report both the total count and the average number (in parenthesis).}
    \vspace{-3ex}
    \label{table:data-stats}
\end{table*}

\subsection{Training}
We design a learning objective that captures both the structural similarity and the content similarity between the ground-truth slides and the generated slides.

\paragraph{Structural similarity.} The series of actions $a^{sec}_j$ and $a^{slide}_k$ determines the \textit{structure} of output slides. To encourage our model to generate slide decks with a similar structure as the ground-truth, we adopt the the cross-entropy loss (CE) and define our structural similarity loss as:
\begin{equation}
    \mathcal{L}_{structure} = \sum\nolimits_j \text{CE}(a^{sec}_j) + \sum\nolimits_k \text{CE}(a^{slide}_k).
\end{equation}

\paragraph{Content Similarity.} We formulate our content similarity loss to capture various aspects of slide generation quality, measuring whether the model 1) selected important sentences and figures from the input document, 2) adequately phrased sentences in the presentation style (e.g., shorter sentences), 3) placed sentences and figures to the right locations on a slide, and 4) put sentences and figures on a slide that are relevant to each other.  We define our content similarity loss to measure each of the four aspects described above:
\begin{equation}
\label{eq:loss-c}
\begin{split}
    \mathcal{L}_{content} & = \sum\nolimits_k \text{CE}(\alpha^{obj}_k) + \sum\nolimits_l \text{CE}(w_l) + \\
    \sum\nolimits_{u,v} & \text{CE}(\delta([E^{txt}_u, E^{fig}_v])) + \sum\nolimits_k \text{MSE}(L_k).
\end{split}
\end{equation}

\textbf{Selection loss ($\alpha^{obj}_k$).} The first term checks whether it selected the ``correct'' objects that also appear in the ground truth. This term is slide-insensitive, i.e., the correct/incorrect inclusion is not affected by which specific slide it appears in.

\textbf{Paraphrasing loss ($w_l$).} The second term measures the quality of paraphrased sentences by comparing the output sentence and the ground-truth sentence word-by-word.

\textbf{Text-Figure matching loss ($\delta([E^{txt}_u, E^{fig}_v])$).} The third term measures the relevance of text and figures appearing in the same slide. We follow the literature on visual-semantic embedding~\cite{kiros14text-image-emb,karpathy14text-image} and learn an additional multimodal projection head $\delta([E^{txt}_u, E^{fig}_v])$ with a sigmoid activation that outputs a relevance score in $[0, 1]$. For positive training pairs, we sample text-figure pairs from a) ground-truth slides and b) paragraph-figure pairs where the figure is mentioned in that paragraph. We randomly construct negative pairs.

\textbf{Layout loss ($L_k$).} The last term measures the quality of slide layout by regressing the predicted bounding box to the ground-truth. While there exist several solutions to bounding box regression~\cite{he15spatial,ren15faster}, we opted for the simple mean squared error (MSE) computed directly over the layout variable $L_k=\{l_k^x, l_k^y, l_k^w, l_k^h\}$.

\paragraph{The Final Loss.} We define our final learning objective as:
\begin{equation}
    \mathcal{L}_{DOC2PPT} = \mathcal{L}_{structure} + \gamma \mathcal{L}_{content}
\end{equation}
where $\gamma$ controls the relative importance between structural and content similarity; we set $\gamma=1$ in our experiments. 

To train our model, we follow the standard teacher-forcing approach~\cite{williams89teacher-forcing} for the sequential prediction and provide the ground-truth results for the past prediction steps, e.g., the next actions $a^{sec}_j$ and $a^{slide}_k$ are based on the ground-truth actions $\tilde{a}^{sec}_{j-1}$ and $\tilde{a}^{slide}_{k-1}$, the next object $\alpha^{obj}_k$ is selected based on the ground-truth object $\tilde{\alpha}^{obj}_{k-1}$, etc.

\subsection{Inference}
The inference procedures during training and test times largely follow the same process, with one exception: At test time, we utilize the multimodal projection head $\delta(\cdot)$ to act as a post-processing tool. That is, once our model generates a slide deck, we remove figures that have relevance scores lower than a threshold $\theta^R$ and add figures with scores higher than a threshold $\theta^A$. We tune the two hyper-parameters $\theta^R$ and $\theta^A$ via cross-validation (we set $\theta^R=0.8$, $\theta^A=0.9$).

\section{Dataset}
We collect pairs of documents and the corresponding slide decks from academic proceedings, focusing on three research communities: computer vision (CVPR, ECCV, BMVC), natural language processing (ACL, NAACL, EMNLP), and machine learning (ICML, NeurIPS, ICLR). Table~\ref{table:data-stats} reports the descriptive statistics of our dataset.

For the training and validation set, we automatically extract text and figures from documents and slides and perform matching to create document-to-slide correspondences. To ensure that our test set is clean and reliable, we use Amazon Mechanical Turk (AMT) and have humans perform image extraction and matching for the entire test set. We provide an overview of our extraction and matching processes; including details of data collection and extraction/matching processes with reliability analyses in the supplementary.

\textbf{Text and Figure Extraction.} For each document $\mathcal{D}$, we extract sections $\mathcal{S}$ and sentences $T^{in}$ using ScienceParse~\cite{url:scienceparse} and figures $\mathcal{F}^{in}$ using PDFFigures~\cite{clark16pdffigures}. For each slide deck $\mathcal{O}$, we extract sentences $\mathcal{T}^{out}$ using Azure OCR~\cite{url:azureocr} and figures $\mathcal{F}^{out}$ using the border following technique~\cite{suzuki85border-following,url:opencv}.

\textbf{Slide Stemming.} Many slides are presented with animations, and this makes $\mathcal{O}$ contain some successive slides that have similar content minus one element on the preceding slide. For simplicity we consider these near-duplicate slides as redundant and remove them by comparing text and image contents of successive slides: if $O_{j+1}$ covers more than 80\% of the content of $O_j$ (per text/visual embeddings) we discard it and keep $O_{j+1}$ as it is deemed more complete. 

\textbf{Slide-Section Matching.} We match slides in a deck to the sections in the corresponding document so that a slide deck is represented as a set of non-overlapping slide groups each with a matching section in the document. To this end, we use RoBERTa~\cite{liu19roberta} to extract embeddings of the text content in each slide and the paragraphs in each section of the document. We assume that a slide deck follows the section order of the corresponding document, and use dynamic programming to find slide-to-section matching based on the cosine similarity between text embeddings. 

\textbf{Sentence Matching.} We match sentences from slides to the corresponding document. We again use RoBERTa to extract embeddings of each sentence in slides and documents, and search for the matching sentence based on the cosine similarity. We limit the search space only within the corresponding sections using the slide-section matching result.

\textbf{Figure Matching.} Lastly, we match figures from slides to those in the corresponding document. We use MobileNet~\cite{howard17mobilenets} to extract visual embeddings of all $I^{in}$ and $I^{out}$ and match them based on the highest cosine similarity. Note that some figures in slides do not appear in the corresponding document (and hence no match). For simplicity, we discard $F^{out}$ if its highest visual embedding similarity is lower than a threshold $\theta^{I}=0.8$.

\begin{table*}[t]
\centering
\footnotesize
    \begin{tabular}[t]{ccccc||cc|ccc|c|c}
        \toprule
        $\:$ & \multicolumn{4}{c||}{Ablation Settings} & \multicolumn{2}{c|}{ROUGE-SL} & \multicolumn{3}{c|}{LC-FS} & \multirow{2}{*}{TFR} & mIoU \\ 
        ~ & Hrch-PT & PAR & TIM & Post Proc. & Ours & w/o SL & Prec. & Rec. & F1 & ~ & (Layout / Template) \\ \midrule
        (a) & \ding{55} & \ding{55} & \ding{55} & \ding{55} & 24.35 & 29.77 & \underline{25.54} & 14.85 & 18.78 & 5.61 & 43.34 / 38.15 \\
        (b) & \ding{51} & \ding{55} & \ding{55} & \ding{55} & 24.93 & 29.68 & 17.48 & \underline{26.26} & 20.99 & 8.58 & \underline{49.16} / 40.94 \\
        (c) & \ding{51} & \ding{51} & \ding{55} & \ding{55} & \underline{27.19} & \underline{32.27} & 17.48 & \underline{26.26} & 20.99 & 9.23 & \underline{49.16} / 40.94 \\
        (d) & \ding{51} & \ding{55} & \ding{51} & \ding{55} & 26.52 & 30.99 & 23.47 & 25.31 & \underline{24.36} & 10.09 & \textbf{50.82} / 42.96 \\
        (e) & \ding{51} & \ding{51} & \ding{51} & \ding{55} & \textbf{29.40} & \textbf{34.27} & 23.47 & 25.31 & \underline{24.36} & \underline{11.82} & \textbf{50.82} / 42.96 \\
        \midrule
        (f) & \ding{51} & \ding{51} & \ding{51} & \ding{51} & \textbf{29.40} & \textbf{34.27} & \textbf{26.36} & \textbf{38.39} & \textbf{31.26} & \textbf{17.49} & $\:\:\:$ - $\:\:\:$ / 46.73 \\
        \bottomrule
    \end{tabular}
    \caption{Overall result of different ablation settings under automatic evaluation metrics ROUGE-SL, LC-FS, TFR, and mIoU.}
    \label{table:overall}
\end{table*}

\begin{table*}[t]
\centering
\footnotesize
    \begin{tabular}[t]{cccccc}
        \toprule
        Train $\downarrow$ \:/\: Test $\rightarrow$ & CV & NLP & ML & All \\
        \midrule
        CV & \textbf{31.2} / \textbf{32.1} / \textbf{19.7} & 24.1 / 21.5 / 5.6 & 24.0 / 25.6 / 11.2 & 24.7 / 29.2 / \underline{15.8} \\
        NLP & 28.8 / 30.0 / 13.4 & \textbf{34.7} / \textbf{30.7} / \textbf{11.8} & 29.2 / 32.7 / 15.3 & \underline{28.9} / 30.9 / 13.6 \\
        ML & 21.1 / 29.2 / 11.6 & 21.1 / 26.6 / 6.6 & \textbf{32.1} / \textbf{36.8} / \textbf{22.8} & 24.9 / \textbf{31.4} / 14.4 \\
        All & \underline{29.2} / \underline{31.2} / \underline{18.6} & \underline{30.0} / \underline{28.8} / \underline{9.7} & \underline{29.4} / \underline{32.9} / \underline{20.6} & \textbf{29.4} / \underline{31.3} / \textbf{17.5} \\
        \bottomrule
    \end{tabular}
    \caption{Topic-aware evaluation results (ROUGE-SL / LC-F1 / TFR) when trained and tested on data from different topics.}
    \vspace{-3ex}
    \label{table:topic}
\end{table*}

\section{Experiments}
\label{sec:experiments}
\docppt is a new task with no established evaluation metrics and baselines. We propose automatic metrics specifically designed for evaluating slide generation methods. We carefully ablate various components of our approach and evaluate them on our proposed metrics. We also perform human evaluation to assess the generation quality.

\subsection{Evaluation Metrics}
\label{sec:eval}
\paragraph{Slide-Level ROUGE (ROUGE-SL).} To measure the quality of text in the generated slides, we adapt the widely-used ROUGE score~\cite{lin14rouge}. Note that ROUGE does not account for the text length in the output, which is problematic for presentation slides (e.g., text in slides are usually shorter). Intuitively, the number of slides in a deck is a good proxy for the overall text length. If too short, too much text will be put on the same slide, making it difficult to read; conversely, if a deck has too many slides, each slide can convey only little information while making the whole presentation lengthy. Therefore, we propose the slide-level ROUGE:
\begin{equation}
    \text{ROUGE-SL} = \text{ROUGE-L} \times e^{\frac{|Q-\tilde{Q}|}{Q}},
\end{equation}
where $Q$ and $\tilde{Q}$ are the number of slides in the generated and the ground-truth slide decks, respectively.

\paragraph{Longest Common Figure Subsequence (LC-FS).} We measure the quality of figures in the output slides by considering both the correctness (whether the figures from the ground-truth deck are included) and the order (whether all the figures are ordered logically -- i.e, in a similar manner to the ground-truth deck). To this end, we use the Longest Common Subsequence (LCS) to compare the list of figures in the output $\{I^{out}_0, I^{out}_1, ...\}$ to the ground-truth $\{\tilde{I}^{out}_0, \tilde{I}^{out}_1, ...\}$ and report precision/recall/F1.

\paragraph{Text-Figure Relevance (TFR).} A good slide deck should put text with relevant figures to make the presentation informative and attractive. We consider text and figures simultaneously and measure their relevance by a modified ROUGE:
\begin{equation}
    \text{TFR} = \frac{1}{M^{in}_F} \sum\nolimits^{M^{in}_F-1}_{i=0} \text{ROUGE-L}(P_{i}, \tilde{P}_{i}),
\end{equation}
where $P_{i}$ and $\tilde{P}_{i}$ are sentences from generated and ground-truth slides that contain $I^{in}_{i}$, respectively. 

\paragraph{Mean Intersection over Union (mIoU).} A good design layout makes it easy to consume information presented in slides. To evaluate the layout quality, we adapt the mean intersection over union (mIoU)~\cite{everingham10pascal} by incorporating the LCS idea with the ground-truth $\tilde{\mathcal{O}}$:
\begin{equation}
    \text{mIoU}(\mathcal{O}, \tilde{\mathcal{O}}) = \frac{1}{N^{out}_{O}} \sum \nolimits^{N^{out}_{O}-1}_{i=0} \text{IoU}(O_i, \tilde{O}_{J_i})
\end{equation}
where $\text{IoU}(O_i, \tilde{O}_j)$ computes the IoU between a set of predicted bounding boxes from slide $i$ and a set of ground-truth bounding boxes from slide and $J_i$. To account for a potential structural mismatch (with missing/extra slides), we find the $J = \{j_0, j_1, ..., j_{N^{out}_{O}-1}\}$ that achieves the maximum mIoU between $\mathcal{O}$ and $\tilde{\mathcal{O}}$ in an increasing order.

\subsection{Implementation Detail}
For the DR, we use a Bi-GRU with 1,024 hidden units and set the MLPs to output 1,024-dimensional embeddings. Each layer of the PT is based on a 256-unit GRU. The PAR is designed as Seq2Seq~\cite{bahdanau15seq2seq} with 512-unit GRU. All the MLPs are two-layer fully-connected networks. We train our network end-to-end using ADAM~\cite{kingma14adam} withlearning rate 3e-4.

\begin{figure*}[t]
\centering
    \includegraphics[width=.9\linewidth]{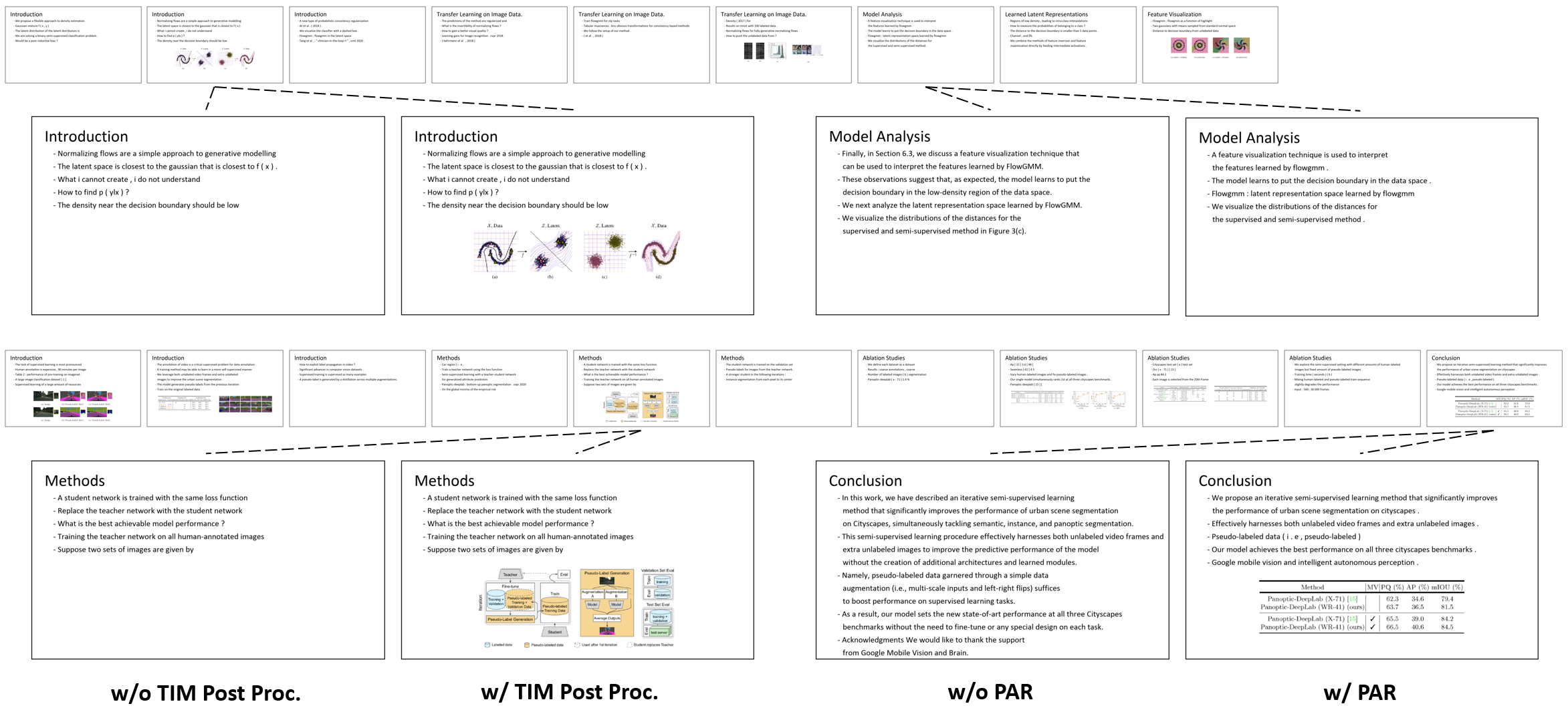}
    \vspace{-1ex}
    \caption{Qualitative examples of the generated slide deck from our model (Paper source: top~\cite{izmailov20flow-gmm} and bottom~\cite{chen20semi}). We provide more results on our project webpage: https://doc2ppt.github.io}
    \vspace{-2ex}
    \label{fig:qual-overall}
\end{figure*}

\begin{figure}[t]
\centering
    \includegraphics[width=.9\linewidth]{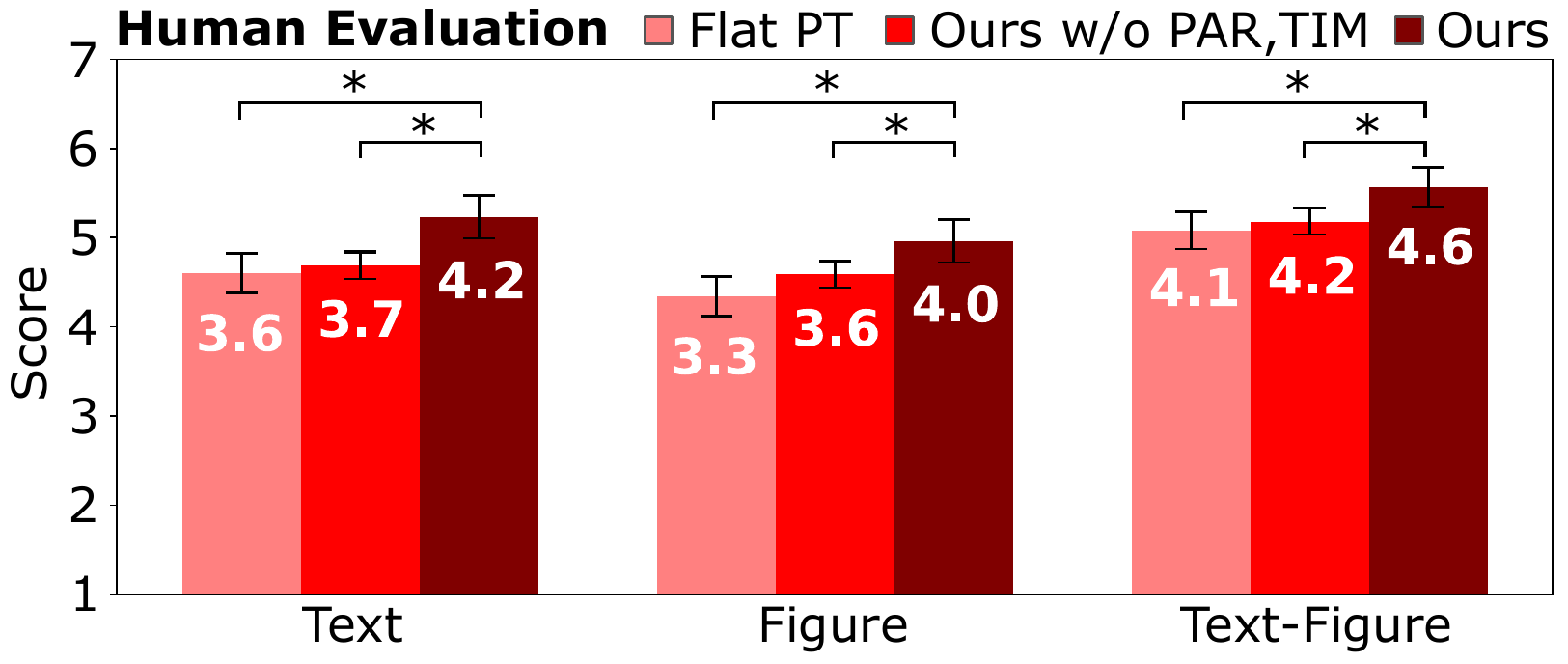}
    \vspace{-1ex}
    \caption{The average scores for how closely the generated slides match the text and figures in the ground-truth slides. And how well the generated text matches the figures in the ground-truth slides. Error bars reflect standard error. Significance tests: two-sample t-test ($p<$0.05.)}
    \vspace{-3ex}
    \label{fig:human}
\end{figure}

\subsection{Results and Discussions}
\paragraph{Is the Hierarchical Modeling Effective?} We define a ``flattened'' version of our PT (flat-PT) by replacing the hierarchical RNN with a vanilla RNN that learns a single shared latent space to model the section-slide-object structure. The flat-PT contains a single GRU and a two-layer MLP with a ternary decision head that learns to predict an action $a_t=\{ \texttt{[NEW\_SECTION]},$ $\texttt{[NEW\_SLIDE]},$ $\texttt{[NEW\_OBJ]} \}$. For a fair comparison, we increase the number of hidden units in the baseline GRU to 512 (ours is 256) so the model capacities are roughly the same between the two. 

First, we compare the structural similarity between the generated and the ground-truth slide decks. For this, we build a list of tokens indicating a section-slide-object structure (e.g., $\texttt{[SEC]}, \texttt{[SLIDE]}, \texttt{[OBJ]}, ..., \texttt{[SLIDE]}, ...$) and compare the lists using the LCS. Our hierarchical approach achieves 64.15\% vs. the flat-PT 51.72\%, suggesting that ours was able to learn the structure better than baseline.

Table~\ref{table:overall} (a) and (b) compare the two models on the four metrics. The results show that ours outperforms flat-PT across all metrics. The flat-PT achieves slightly better performance on ROUGE-SL without the slide-length term (w/o SL), which is the same as ROUGE-L. This suggests that ours generates a slide structure more similar to the ground-truth.

\paragraph{A Deeper Look into the Content Similarity Loss.} We ablate different terms in the content similarity loss (Eq.~\ref{eq:loss-c}) to understand their individual effectiveness in Table~\ref{table:overall}.

\textbf{PAR.} The paraphrasing loss improves text quality in slides; see the ROUGE-SL scores of (b) vs. (c), and (d) vs. (e). It also improves the TFR metric because any improvement in text quality will benefit text-figure relevance.

\textbf{TIM.} The text-figure matching loss improves the figure quality; see (b) vs. (d) and (c) vs. (e). It particularly improves LC-FS precision with a moderate drop in recall rate, indicating the model added more correct figures. TIM also improves ROUGE-SL because it helps constrain the multimodal embedding space, resulting in better selection of text.

\paragraph{Figure Post-Processing.} At test time, we leverage the multimodal projection head $\delta(\cdot)$ as a post-processing module to add missing figures and/or remove unnecessary ones. Table~\ref{table:overall} (f) shows this post-processing further improves the two image-related metrics, LC-FS and TFR. For simplicity, we add figures following equally fitting in template-based design instead of using OP to predict its location. 

\paragraph{Layout Prediction vs. Template.} The OP predicts the layout to decide where and how to put the extracted objects. We compare this with a template-based approach, which selects the current section title as the slide title and puts sentences and figures in the body line-by-line. For those extracted figures, they will equally fit (with the same width) in the remaining space under the main content. The result shows that the predicted-based layout, which directly learns from the layout loss, can bring out higher mIoU with the groundtruth. And in the aspect of the visualization, the template-based design can make the generated slide deck more consistent.

\paragraph{Topic-Aware Evaluation.} We evaluate performance in a topic-dependent and independent fashion. To do this, we train and test our model on data from each of the three research communities (CV, NLP, and ML). Table~\ref{table:topic} shows that models trained and tested within each topic performs the best (not surprisingly), and that models trained on data from all topics achieves the second best performance, showing generalization to different topic areas. Training on NLP data, despite being the smallest among the three, seems to generalize well to other topics on the text metric, achieving the second best on ROUGE-SL (28.9). Training on CV data provides the second highest performance on the text-figure metric TFR (15.8), and training on ML achieves the highest figure extraction performance (LC-FS F1 of 31.4).

\paragraph{Human Evaluation.} We conduct a user study to assess the perceived quality of generates slides. To make the task easy to complete, we sample 200 sections from 50 documents and create 600 pairs of ground-truth and generate slides. We prepare four slide decks per document: the ground-truth deck, and the ones generated by the flat PT (Table~\ref{table:overall} (a)), by ours without PAR and TIM (b), and by our final model (f).

We recruited three AMT Master Workers for each task (HIT). The workers were shown the slides from the ground-truth deck (DECK A) and one of the methods (DECK B). The workers were then asked to answer three questions: \texttt{Q1}. Looking only at the TEXT on the slides, how similar is the content on the slides in DECK A to the content on the slides in DECK B?; \texttt{Q2}. How well do the figure(s)/tables(s) in DECK A match the text or figures/tables in DECK B?; \texttt{Q3}. How well do the figure(s)/table(s) in DECK A match the TEXT in DECK B? The responses were all on a scale of 1 (not similar at all) to 7 (very similar). 
Fig.~\ref{fig:human} shows the average scores for each method. The average rating for our approach was significantly greater for all three questions compared to the other two methods. There was no significant difference between the ratings for the other two methods.

\paragraph{Qualitative Results.} Fig.~\ref{fig:qual-overall} illustrates two qualitative examples of the slide deck generated by our model with the template-based layout generation approach. With the post-processing, TIM can add the related figure into the slide and make it more informative. PAR helps create a better presentation by paraphrasing the sentences into bullet point form. 

\section{Conclusion}
We present a novel task and approach for generating slides from documents. This is a challenging multimodal task that involves understanding and summarizing documents containing text and figures and structuring it into a presentation form. We release a large set of 5,873 paired documents and slide decks, and provide evaluation metrics with our results. We hope our work will help advance the state-of-the-art in vision-and-language understanding.

\onecolumn \appendix
\section{Details of the Data Processing Steps}
\begin{figure}[!ht]
\centering
    \includegraphics[width=.95\linewidth]{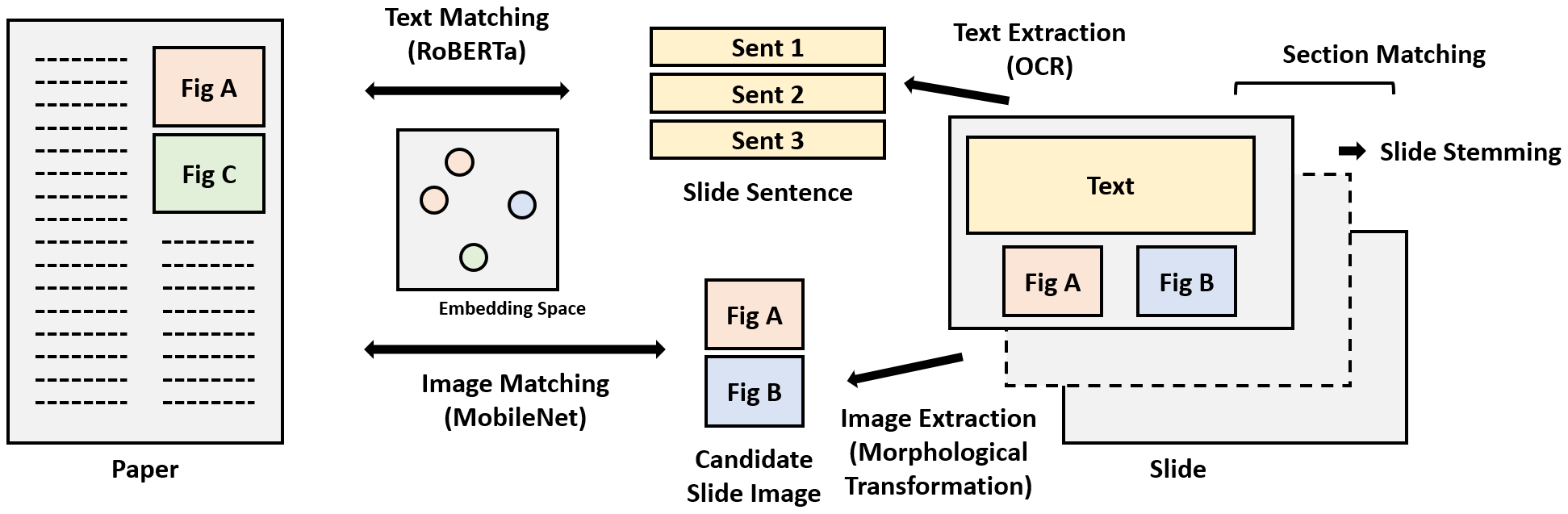}
    \vspace{-1ex}
    \caption{\textbf{Data processing pipeline.} We automatically extract text/figures and match them between documents and slide decks.}
    \vspace{-2ex}
    \label{fig:preprocessing}
\end{figure}

Section 4 in our main paper explains how we construct our \docppt dataset. Here we provide the details of the process and demonstrate the accuracy of the various extraction/matching processes. Fig.~\ref{fig:preprocessing} illustrates the details of the data processing pipeline that were omitted in the main paper. To evaluate how reliable the various steps in our pipeline are, we manually labeled 100 slide decks (randomly sampled from the validation split) and used them for evaluation.

\begin{figure}[!ht]
\centering
    \includegraphics[width=.95\linewidth]{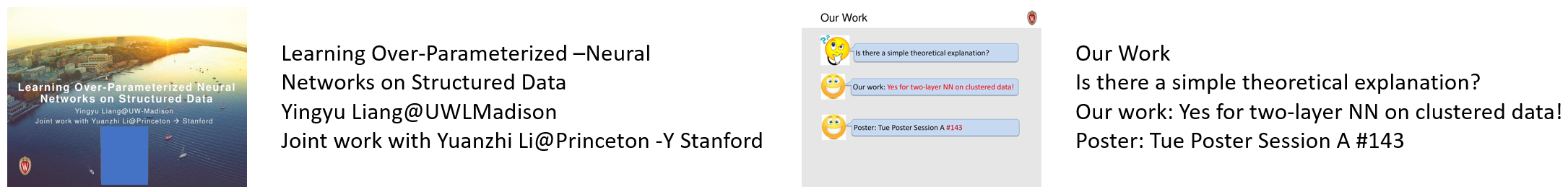}
    \vspace{-1ex}
    \caption{\textbf{Text Extraction from Slide Deck.} We use Azure OCR~\cite{url:azureocr} to extract sentences from slides.}
    \vspace{-2ex}
    \label{fig:ocr}
\end{figure}

\paragraph{Text Extraction}
Fig.~\ref{fig:ocr} shows examples of the extracted slide sentences obtained using Azure OCR~\cite{url:azureocr}. The slides are shown on left and the extracted text is on the right. Notice that the OCR results are quite reliable as slides contain text.

\begin{figure}[!ht]
\centering
    \includegraphics[width=.95\linewidth]{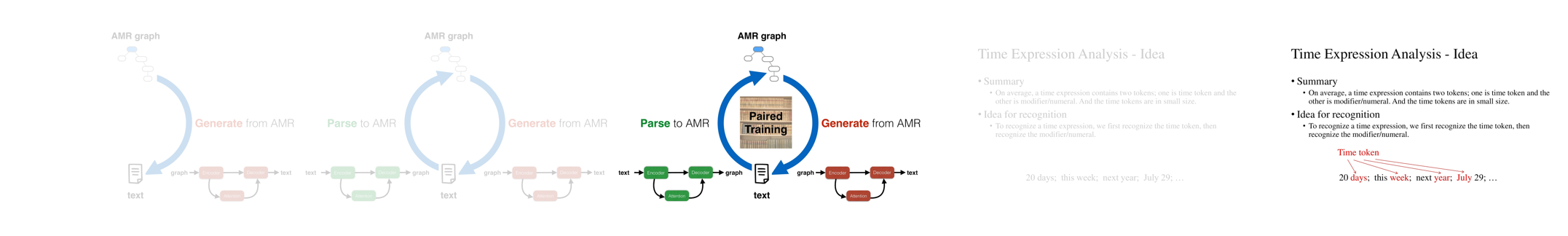}
    \vspace{-1ex}
    \caption{\textbf{Slide Stemming.} The ghosted/opaque slides are seen as redundant and will be removed by the stemming process. This helps simplify our dataset.}
    \vspace{-2ex}
    \label{fig:stemming}
\end{figure}

\paragraph{Slide Stemming}
Fig.~\ref{fig:stemming} illustrates the slide stemming process. If a slide has a preceding slide with 80\% or greater overlap in content, we consider the preceding slide as redundant and remove it. The slides which are opaque (ghosted) are examples of slides that would be removed (they often exist because of animations that sequentially add elements to a slide - e.g., bullet points appearing - thus we just keep the final slide in the sequence to simplify the dataset). Our slide stemming step is 93\% accurate based on the human annotated validation set.

\begin{figure}[!ht]
\centering
    \includegraphics[width=.95\linewidth]{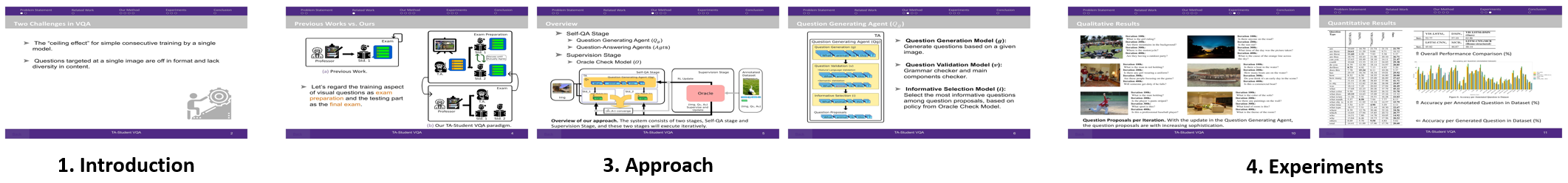}
    \vspace{-1ex}
    \caption{\textbf{Slide-Section Matching.} We match slides to the corresponding sections in the document so that a slide deck is represented as a set of non-overlapping section groups.}
    \label{fig:section-matching}
\end{figure}

\paragraph{Slide-Section Matching}
Fig.~\ref{fig:section-matching} presents an example of slide-section matching. We adopt RoBERTa \cite{liu19roberta} to extract embeddings of the text in slides and sections in the document (paper). Specifically, we find slide-to-section matching based on the cosine similarity between text embeddings. Slides are matched with the section with the highest cosine similarity and our slide-section matching has 82\% accuracy. 

\paragraph{Sentence Matching}
Table~\ref{table:sentence-matching} shows examples of matching sentences between the paper and the slide. We again use RoBERTa to search for the matching sentence based on the cosine similarity and build the linking for the extractive summarization.
\begin{table}[!ht]
\centering
\footnotesize
    \begin{tabular}[t]{cc}
        \toprule
        Paper Sentence & Slide Sentence \\
        \midrule
        The Pima Indians Diabetes data set & This data set contains 768 \\
        contains information about 768 diabetes & diabetes patients, \\
        patients, recording features like glucose & recording features like \\
        blood, pressure, age, and skin thickness & glucose, blood \\
        \midrule
        Finally, can the idea of proportionality  & Can fairness as \\ 
        as a group fairness concept be adapted & proportionality be \\
        for supervised learning tasks & adapted for supervised \\
        like classification and regression \\
        \bottomrule
    \end{tabular}
    \caption{\textbf{Sentence Matching.} The example of matching sentences from the slide to the paper.}
    \label{table:sentence-matching}
    \vspace{-3ex}
\end{table}

\begin{figure}[!ht]
\centering
    \begin{minipage}{.45\linewidth}
        \centering
        \includegraphics[width=\linewidth]{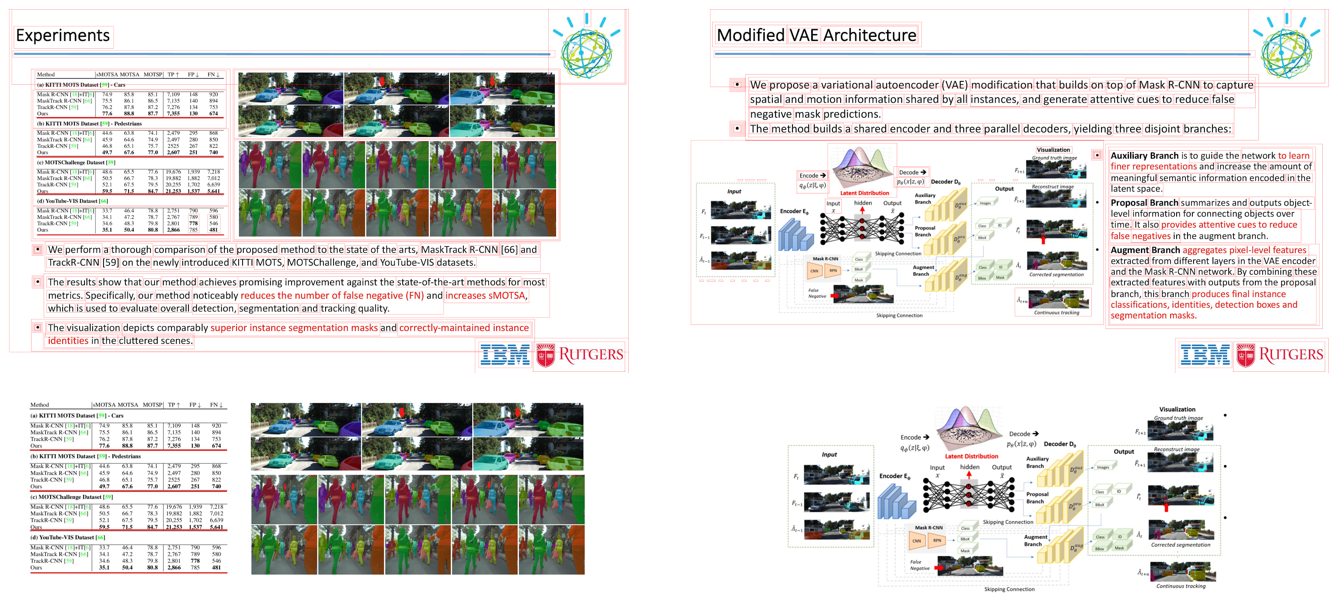}
        \vspace{-2ex}
        \caption{\textbf{Figure Matching.} The lower figures are those matched from the paper using the cosine similarity and features from MobileNet \cite{howard17mobilenets}.}
        \label{fig:figure-matching}
    \end{minipage}~~~~~
    \begin{minipage}{.45\linewidth}
        \centering
        \includegraphics[width=\linewidth]{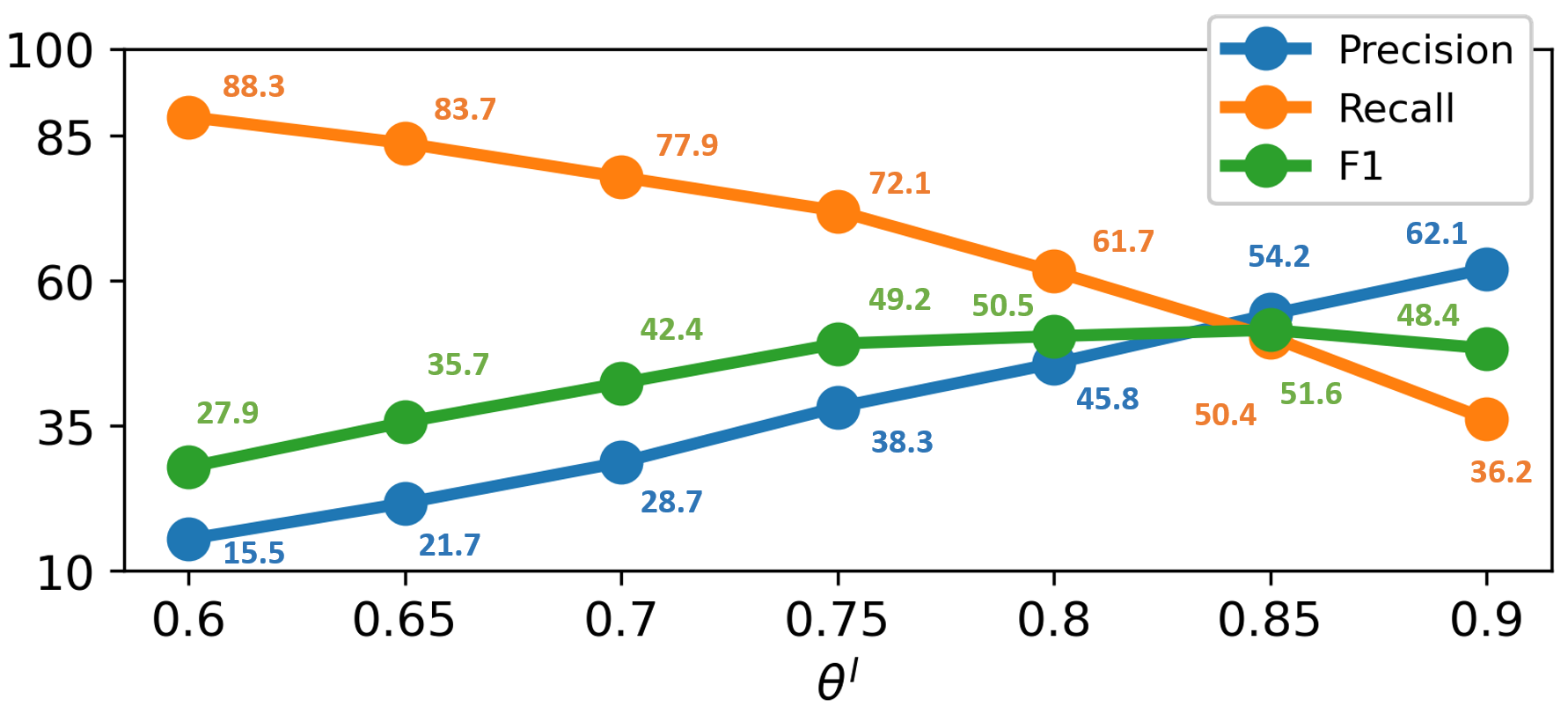}
        \vspace{-2ex}
        \caption{\textbf{Figure Matching under Different $\theta^I$.} Precision, recall, and F1 are evaluated using the human-labeled testing set.}
        \label{fig:threshold-figure-matching}
    \end{minipage}
    \vspace{-2ex}
\end{figure}

\begin{figure}[!ht]
\centering
    \includegraphics[width=.95\linewidth]{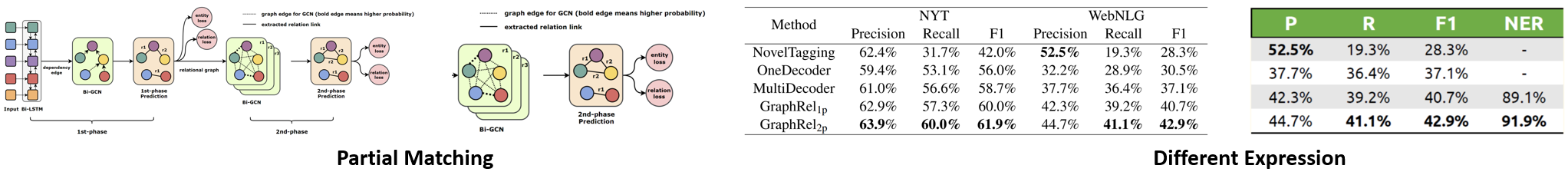}
    \vspace{-1ex}
    \caption{\textbf{Partial Matching and Different Expression.} The examples where the figure matching performs poorly.}
    \vspace{-2ex}
    \label{fig:fail-figure-matching}
\end{figure}

\paragraph{Figure Matching}
Fig.~\ref{fig:figure-matching} illustrates examples of figures/tables that were matched with a particular slide. We apply morphological transformation \cite{url:opencv} and border following \cite{suzuki85border-following} to extract possible slide figures. We then match them with figures in the paper using the visual embedding from MobileNet \cite{howard17mobilenets}; if the cosine similarity is larger than the threshold $\theta_I$. Fig.~\ref{fig:threshold-figure-matching} presents the precision, recall, and F1, which are evaluated from human-labeled test set. The x-axis represents different values of threshold $\theta^I$ considered when comparing the cosine similarity of the visual embedding. When $\theta^I$ is lower, more figures from the paper will be included, which increases recall but negatively impacts precision; in contrast, a higher $\theta^I$ results in greater precision but lower recall. Fig.~\ref{fig:fail-figure-matching} shows examples where the figure matching performs poorly. There are two cases: 1) partial figure matches where a figure has had elements added or removed, and 2) different versions of a figure where the meaning might be similar but the images do not match. These cases make matching difficult, because based on the visual embedding they may not be very similar.

\begin{figure}[!ht]
\centering
    \begin{minipage}{.48\linewidth}
        \centering
        \includegraphics[width=\linewidth]{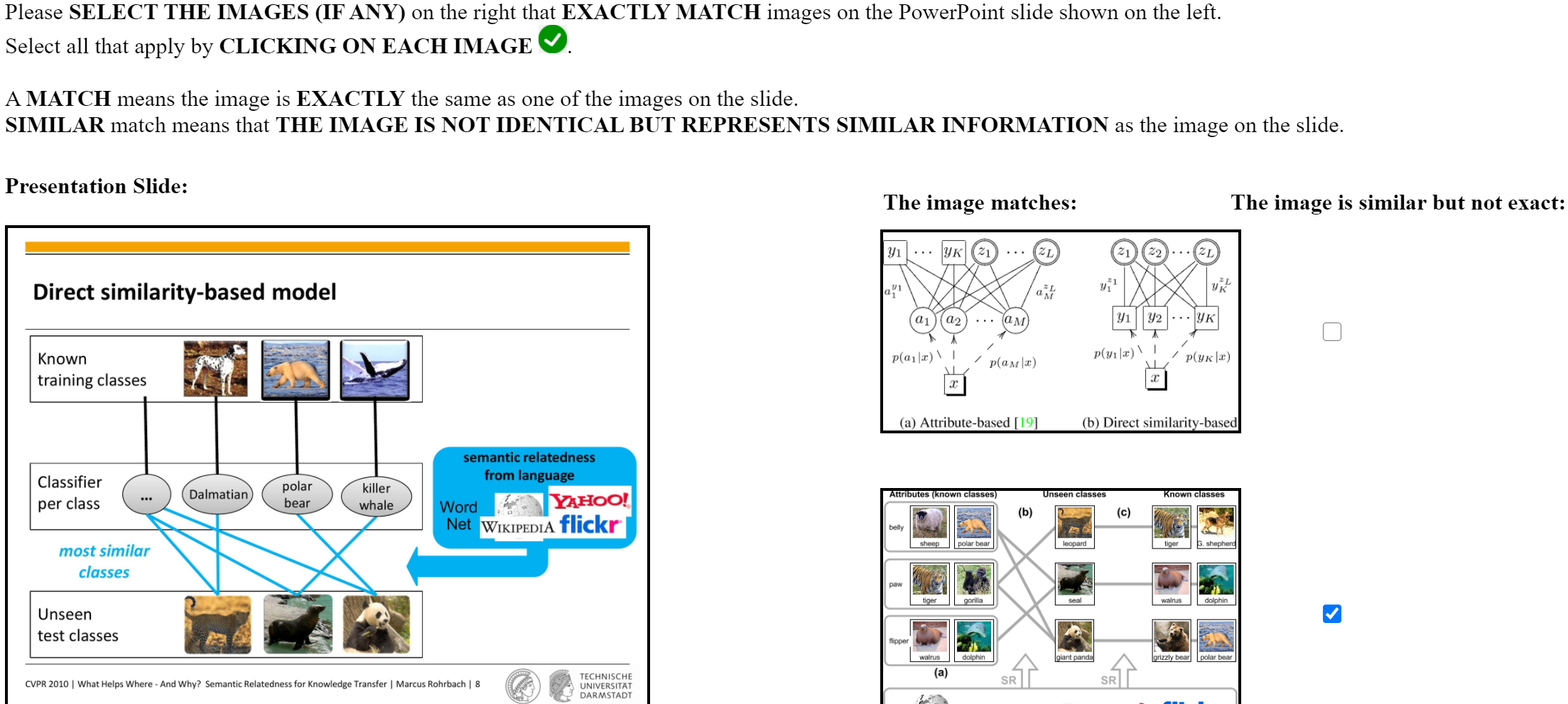}
        \vspace{-2ex}
        \caption{\textbf{Interface of Figure Matching Labeling.} The annotator label figures either as match or as similar but not exact match.}
        \label{fig:human-figure-matching}
    \end{minipage}~~~~~
    \begin{minipage}{.4\linewidth}
        \centering
        \includegraphics[width=\linewidth]{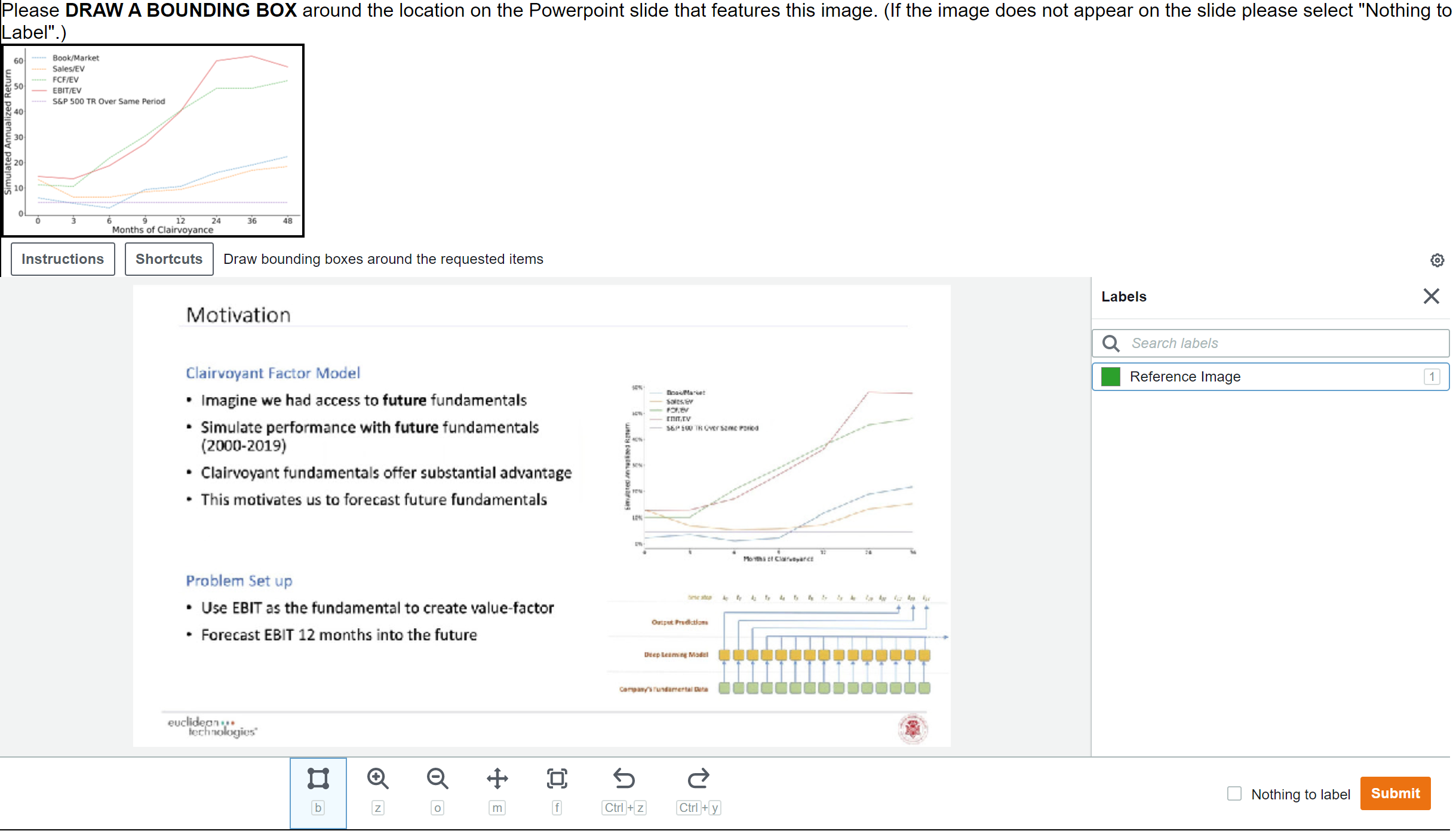}
        \vspace{-2ex}
        \caption{\textbf{Interface of Bounding Box Labeling.} The annotator is asked to draw a bounding box around the region of the slide where the candidate figure appeared.}
        \label{fig:human-bounding-box}
    \end{minipage}
    \vspace{-2ex}
\end{figure}

\paragraph{Human Labeling}
To ensure that our test set is clean and reliable, we use Amazon Mechanical Turk (AMT) and have humans perform image extraction and matching for the entire testing set. Fig.~\ref{fig:human-figure-matching} shows a screenshot of the MTurk HIT for labeling figure matches within each slide. The slide is shown on the left and figures from the document (paper) were shown on the right.  The human annotators can label each figure either as a match (by clicking on the image) or as similar but not an exact match (by ticking the checkbox next to the image). Fig.~\ref{fig:human-bounding-box} shows a screenshot of the MTurk HIT for labeling the bounding box around the image on a slide. The candidate figure is shown above and the human annotator is asked to draw a bounding box around the region of the slide where it appeared. We perform figure-slide matching (see above) before bounding box labeling as this produced the best quality annotations (bounding box labeling is not necessary if the image isn't on the slide at all). For the human-labeled testing set, a slide deck contains on average 2.3 images that are excerpted from the corresponding paper. Please note that since people tend to adopt more new figures or different figures in a slide deck for computer vision (CV) field, the average number of excerpted figure is lower (1.7).

\section{Settings of Approach}
\begin{figure}[!ht]
\centering
    \begin{minipage}{.4\linewidth}
    \small
        \centering
        \begin{tabular}[t]{ccccc}
            \toprule
            \multicolumn{3}{c}{Ablation Settings} & \multicolumn{2}{c}{Rouge-L} \\
            Hrch-PT & TIM & PAR & w/o $h^{obj}$ & w/ $h^{obj}$ \\
            \midrule
            \ding{51} & \ding{55} & \ding{55} & \multicolumn{2}{c}{29.68} \\
            \ding{51} & \ding{55} & \ding{51} & 31.95 & \textbf{32.27} \\
            \ding{51} & \ding{51} & \ding{55} & \multicolumn{2}{c}{30.99} \\
            \ding{51} & \ding{51} & \ding{51} & 33.05 & \textbf{34.27} \\
            \bottomrule
            \\
        \end{tabular}
        \vspace{-2ex}
        \captionof{table}{\textbf{Considering $h^{pbj}$ in PAR.} The Rouge-L score of with and without $h^{obj}$ in paraphrasing module.}
        \label{table:par-h}
    \end{minipage}~~~~~
    \begin{minipage}{.59\linewidth}
        \centering
        \includegraphics[width=\linewidth]{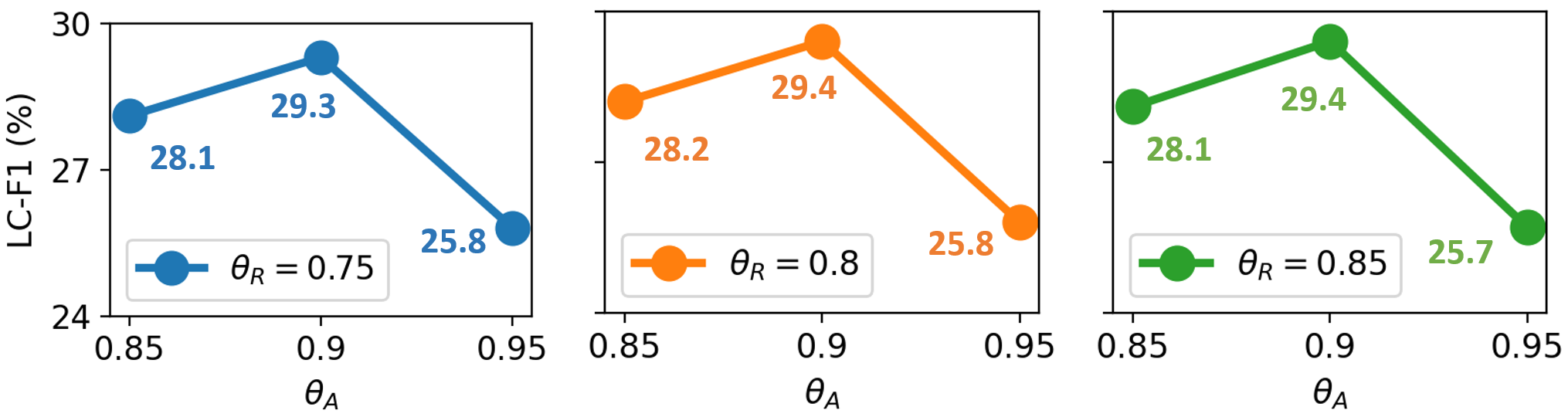}
        \vspace{-2ex}
        \caption{\textbf{Post-Processing under different $\theta^R$ and $\theta^A$.} We tune the $\theta^R$ and $\theta^A$ for the post-processing based on the LC-F1 on the validation set.}
        \label{fig:threshold-postproc}
    \end{minipage}
    \vspace{-2ex}
\end{figure}

\paragraph{The importance of $h^{obj}$ in Paraphrasing Module}
Table~\ref{table:par-h} presents the Rouge-L of paraphrasing module (PAR) with or without using the object state $h^{obj}$. The results show that the text quality improves in all cases if we apply PAR. Also, using $h^{obj}$ benefits more (w/ 32.27 vs w/o 31.95). This is because $h^{obj}$ provides contextual information, which helps PAR generate a paraphrased sentence more relevant to the content in the document.

\paragraph{Sensitivity of $\theta^R$ and $\theta^A$ in Post-Processing}
During the post-processing, we remove figures deemed irrelevant by $\theta^R$ and add ones if considered highly relevant based on $\theta^A$. To achieve the best result, we tune our $\theta^R$ and $\theta^A$ on the 100 labeled validation set. Fig.~\ref{fig:threshold-postproc} shows that $\theta^R=0.8$ and $\theta^A=0.9$ achieves the highest LC-F1.

\section{Inference Flow}
\begin{figure}[!ht]
\centering
    \includegraphics[width=.95\linewidth]{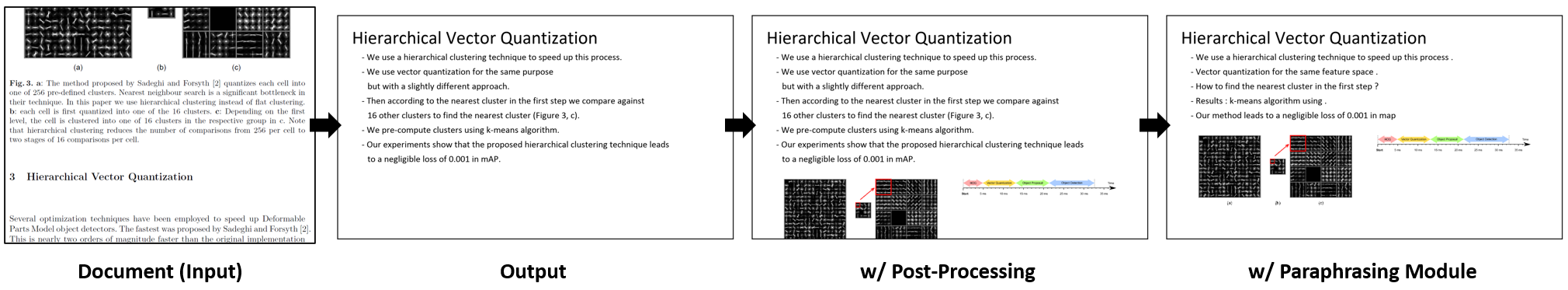}
    \vspace{-1ex}
    \caption{\textbf{Inference Flow.}}
    \label{fig:flow}
\end{figure}

Fig.~\ref{fig:flow} illustrates the inference flow of the proposed approach. Given an academic paper as input, we will first have a generated slide deck from our model. During the post-processing, there is an opportunity to remove unrelated figures and add related ones, and make the slide deck more attractive. By paraphrasing, PAR can further help transform sentences into slide-style.

\section{Human Evaluation}
\begin{figure}[!ht]
\centering
    \begin{minipage}{.4\linewidth}
        \centering
        \includegraphics[width=\linewidth]{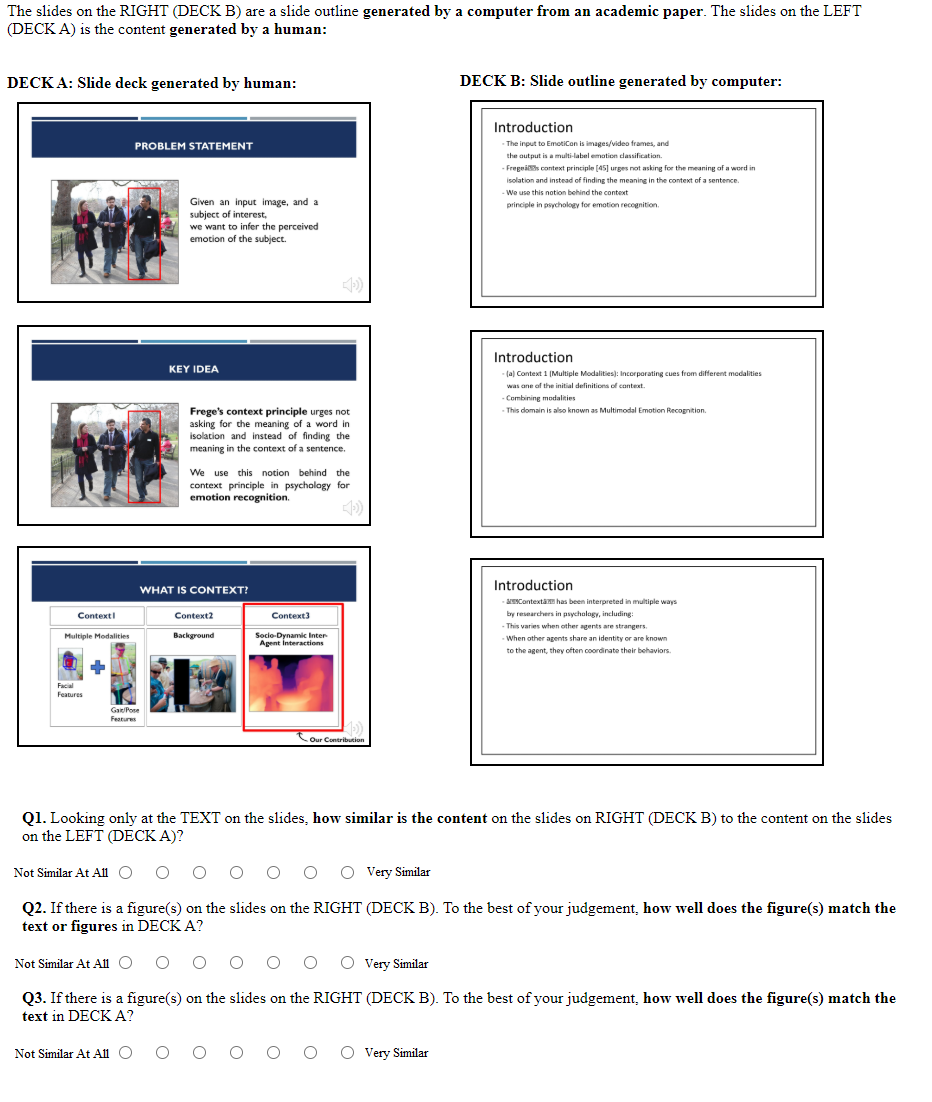}
        \vspace{-2ex}
        \caption{\textbf{Interface of Human Evaluation.} The annotator is asked three questions on aspects of the text quality, the figure extraction, and the text-figure relevance.}
        \label{fig:human-evaluation}
    \end{minipage}~~~~~
    \begin{minipage}{.5\linewidth}
        \centering
        \includegraphics[width=\linewidth]{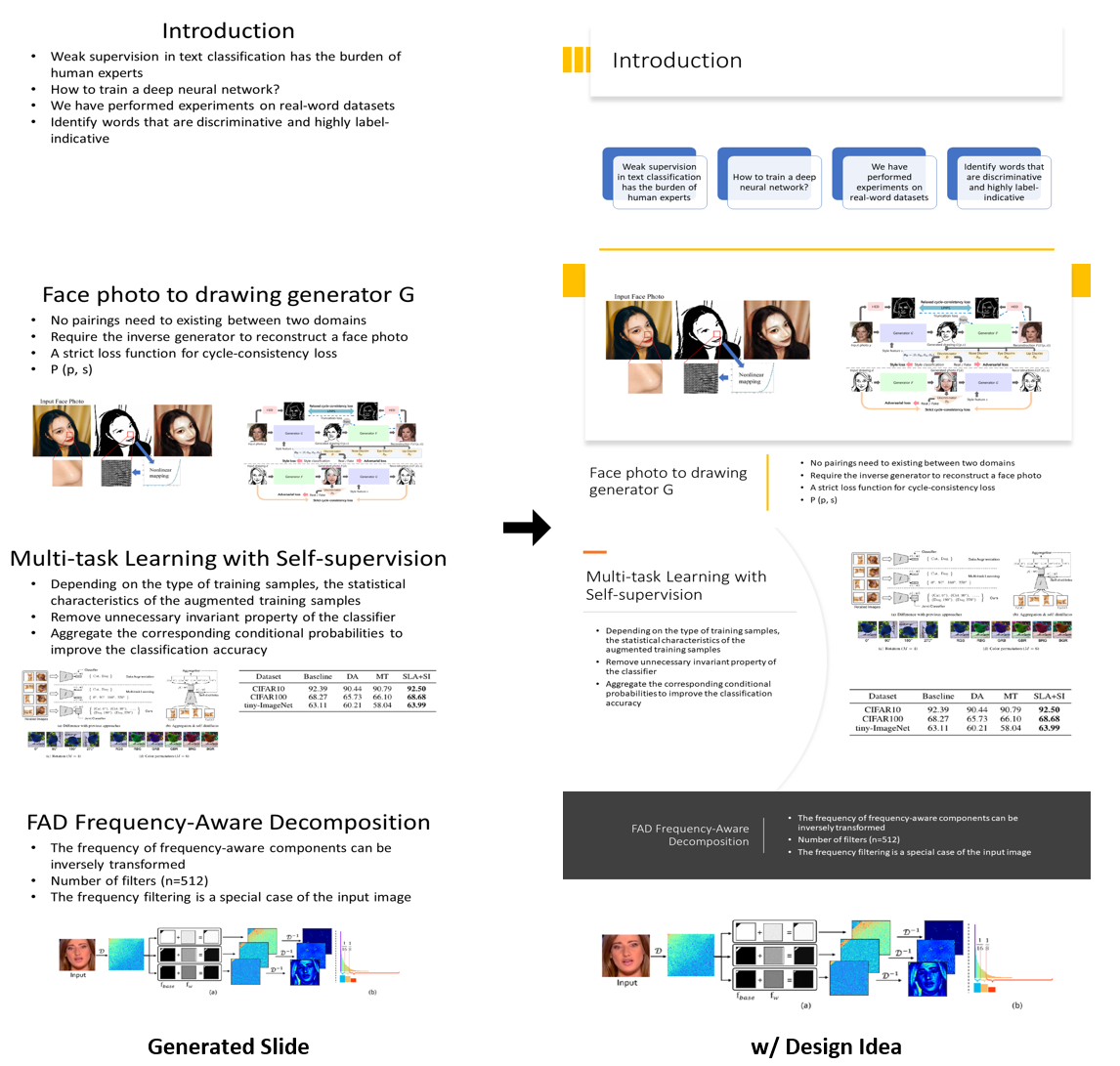}
        \vspace{-2ex}
        \caption{\textbf{Applying PowerPoint Design Ideas.} By applying the design ideas feature \cite{url:design-ideas} provided from Microsoft PowerPoint, we can make the generated template-based slide deck more professional and more attractive for the presentation.}
        \label{fig:design-idea}
    \end{minipage}
    \vspace{-2ex}
\end{figure}

Fig.~\ref{fig:human-evaluation} shows a screenshot of the human rating task for evaluating the quality of the generated slides. The ground-truth slide deck was shown (left) alongside the generated slides (right).  The human annotators were asked three questions. 1) How similar the text on slide DECK A was to the text on slide DECK B. 2) How similar the figures on slide DECK A was to the figures on slide DECK B - the could also indicate that no figures were present.  3) How similar the figures in DECK B were to the text in DECK A - again they could indicate that no figures were present if that was the case.

\section{Qualitative Examples}
\begin{figure}[!ht]
\centering
    \includegraphics[width=\linewidth]{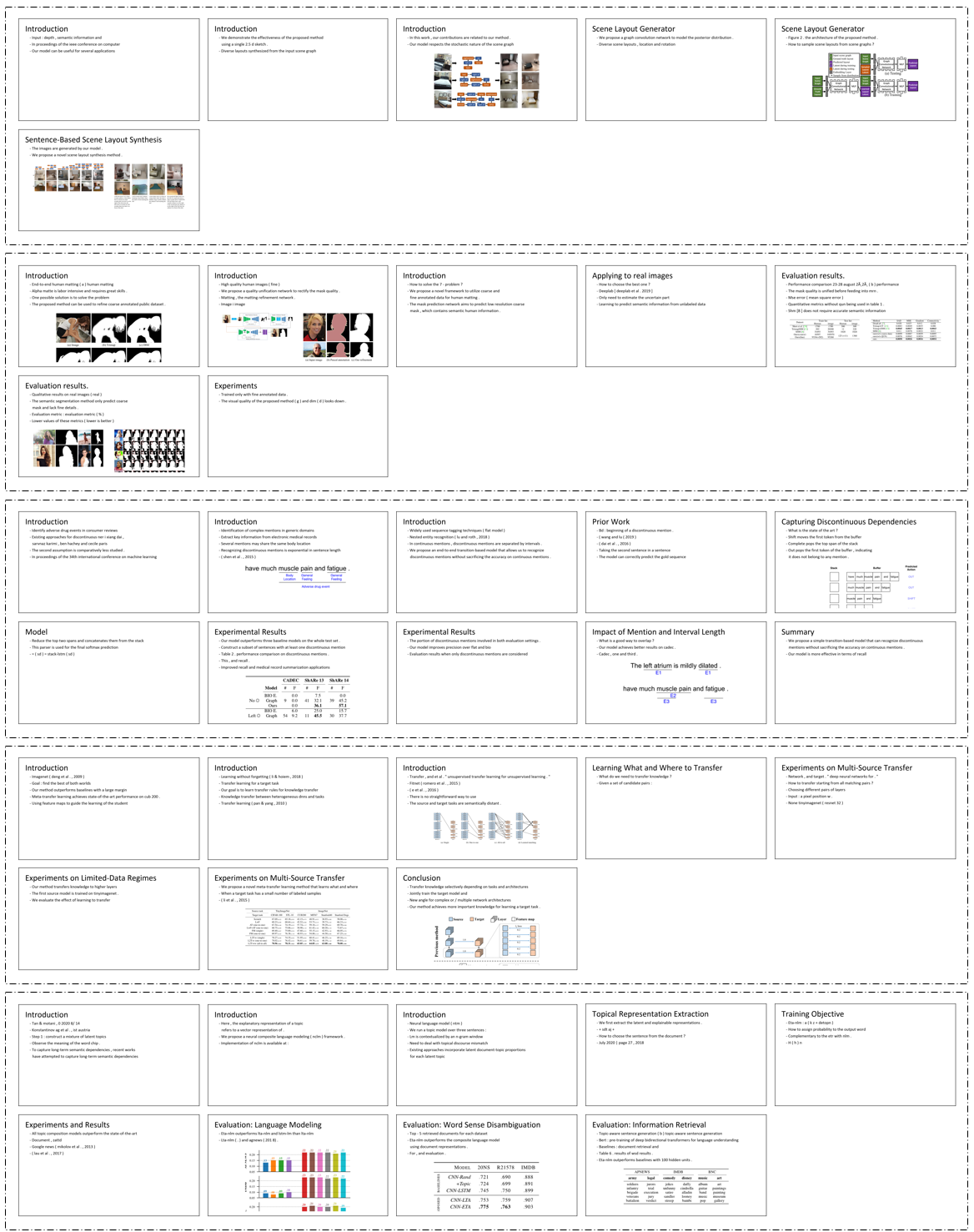}
    \vspace{-2ex}
    \caption{\textbf{Qualitative examples.} (from top to bottom: \cite{luo20e2e-scene}, \cite{liu20human-matting}, \cite{dai20ner}, \cite{jang19learn-transfer}, and \cite{chaudhary20topic-nlu}) Please visit https://doc2ppt.github.io for more generated slide decks.}
    \label{fig:qual-big}
\end{figure}

Fig.~\ref{fig:qual-big} demonstrates generated slide decks from our approach. We provide more results, including failure cases, on our project webpage: https://doc2ppt.github.io.

\paragraph{Applying PowerPoint Design Ideas} As we discussed in the main paper, the output of our method can be used as a draft slide deck for humans to build upon. We provide one such application scenario of our approach. When the slide decks are generated based on a template, the content are all in a fixed size and in the fix position. To make the output more attractive, we can apply off-the-shelf tools such as Microsoft PowerPoint Design Ideas~\cite{url:design-ideas}  which can automatically produce a layout for the given texts and figures. As shown in Fig.~\ref{fig:design-idea}, the generated decks are more professional looking.

\clearpage
\bibliography{aaai22}

\end{document}